\documentclass[11pt,a4paper]{article}

\usepackage[utf8]{inputenc}
\usepackage[T1]{fontenc}
\usepackage[english]{babel}
\usepackage{graphicx}
\usepackage{xcolor}
\usepackage{amsmath,amssymb}
\usepackage{bm}
\usepackage{caption}
\usepackage{cite}
\usepackage{xurl}
\usepackage{bbm}
\usepackage[colorlinks=true,urlcolor=blue,citecolor=blue,linkcolor=blue]{hyperref}

\hypersetup{
  pdftitle={Mathematical Analysis of Image Matching Techniques},
  pdfauthor={O. Samoilenko},
  pdfsubject={https://doi.org/10.37069/1683-4720-2025-39-8},
  pdfkeywords={image matching, feature detection, computer vision, pattern recognition}
}

\title{Mathematical Analysis of Image Matching Techniques\\[2mm]
  \small\normalfont{Published in \emph{Proceedings of the Institute of Applied Mathematics and Mechanics NAS of Ukraine}, 39 (2025).
  \url{https://doi.org/10.37069/1683-4720-2025-39-8}}}

\author{O. Samoilenko%
  \thanks{The authors confirm that there is no conflict of interest and acknowledges financial support by
  the Simons Foundation grant (SFI-PD-Ukraine-00014586, O.S.) and the National Academy of Sciences of Ukraine under the
  budget programme 0125U000299.}\\
  \small Institute of Mathematics, National Academy of Sciences of Ukraine, Kyiv, Ukraine\\
  \small \texttt{oleh.samoilenko@imath.kiev.ua}}

\date{}

\begin{document}

\maketitle

\begin{abstract}
Image matching is a fundamental problem in Computer Vision with direct applications in robotics, remote sensing and geospatial data analysis. 
The present paper reports an analytical and experimental evaluation of classical local feature-based image matching algorithms in the context of satellite imagery, 
specifically, the Scale-Invariant Feature Transform (SIFT) and the Oriented FAST and Rotated BRIEF (ORB) that is built on the Features from Accelerated Segment Test (FAST) keypoint detector and the Binary Robust Independent Elementary Features (BRIEF) descriptor.
Each method is decomposed into a common matching pipeline consisting of the keypoint detection, local descriptor extraction, descriptor matching, and geometric verification.
Formally, an image with \( c \) color channels is modeled as a function \( I: \mathbb{R}^2 \rightarrow \mathbb{R}^c\).
Keypoints (features) are locally distinctive points in the image, defined as spatial locations \( \mathbf{x} \in \mathbb{R}^2 \).
Each keypoint is associated with a compact descriptor vector \( f(\mathbf{x}) \in \mathbb{R}^D \), where $D$ denotes the descriptor dimensionality. 
The descriptor matching stage attempts to establish a correspondence \( f(\mathbf{x}_i) \approx f(\mathbf{x}'_i) \) between sets of points according to a predefined distance metric,
such that the pair \( (\mathbf{x}_i,\mathbf{x}'_i) \) represents the same physical location.
To ensure geometric consistency, we incorporate a verification employing the Random Sample Consensus (RANSAC) algorithm, where a ho\-mog\-ra\-phy matrix \( \mathbf{H} \in R^{3 \times 3} \) is estimated to model a projective transformation between the images.
A match is denoted as an inlier (or valid match) if it satisfies the condition \( \| \mathbf{x}'_i - \mathbf{H} \mathbf{x}_i \| < \epsilon \) for a predefined threshold $\epsilon$.
The proportion of correspondences satisfying this constraint is referred to as the Inlier Ratio, is used as a measure of matching confidence and serves as the primary metric in the evaluation.
The study utilizes a manually constructed dataset of satellite images.
The dataset includes GPS-annotated map tiles with intentional partial overlaps between adjacent images enabling a reliable evaluation via a pairwise image matching.
We examine the impact of varying the number of extracted keypoints on the resulting Inlier Ratio.
\vspace{1mm}
\vspace{1mm}
\end{abstract}

\section{Introduction}

% Context and Motivation, Problem statement?
Image matching is a fundamental problem in Computer Vision. %with applications image retrieval.
A particularly challenging and important subproblem involves matching the satellite images due to repet\-i\-tive patterns (e.g. grid-like street layouts, rows of buildings, or agricultural fields) which can lead to a high number of ambiguous matches, illumination and weather variations, temporal dynamics (e.g. new construction or demolition), limited texture in certain terrains (e.g. deserts, forests, water bodies) where distinctive keypoints may be sparse or nonexistent. 

The motivation for the matching map images arises from numerous real-world ap\-pli\-ca\-tions in robotics, remote sensing, and visual geolocatization.
Given the visual variability of the map data, robust matching techniques should be designed to handle the affine distortions, noise, and limited texture information.
This stimulates the in\-ves\-ti\-ga\-tion and evaluation of the feature matching methods, specifically for cartographic imagery.
Understanding the strengths and limitations of these methods in the context of the map matching can significantly contribute to improvements in the geospatial data processing and automated map analysis systems.

\begin{figure}[ht]
    \centering

    \begin{minipage}[b]{0.32\textwidth}
        \includegraphics[width=\textwidth]{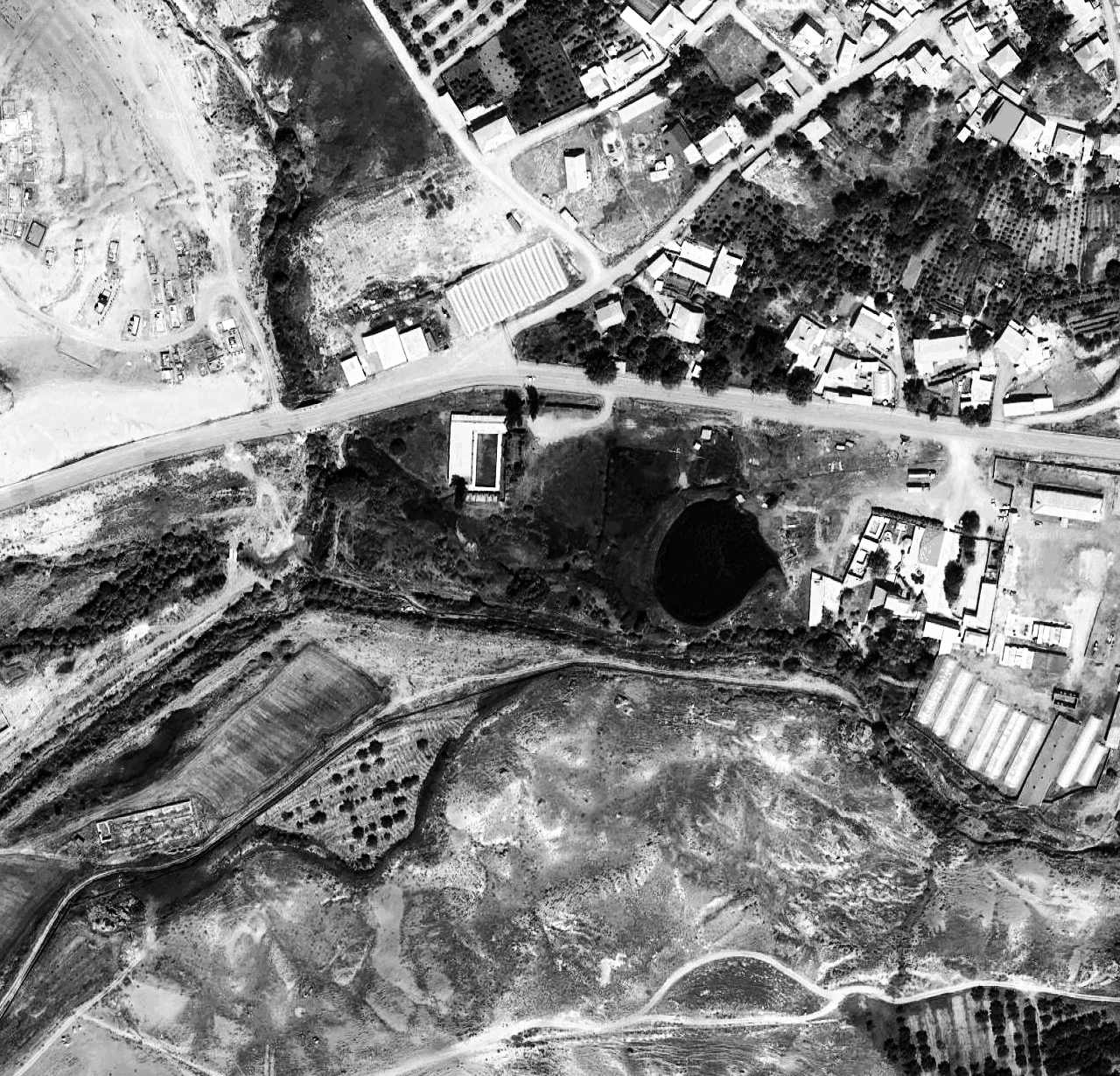}
    \end{minipage}
    \begin{minipage}[b]{0.32\textwidth}
        \includegraphics[width=\textwidth]{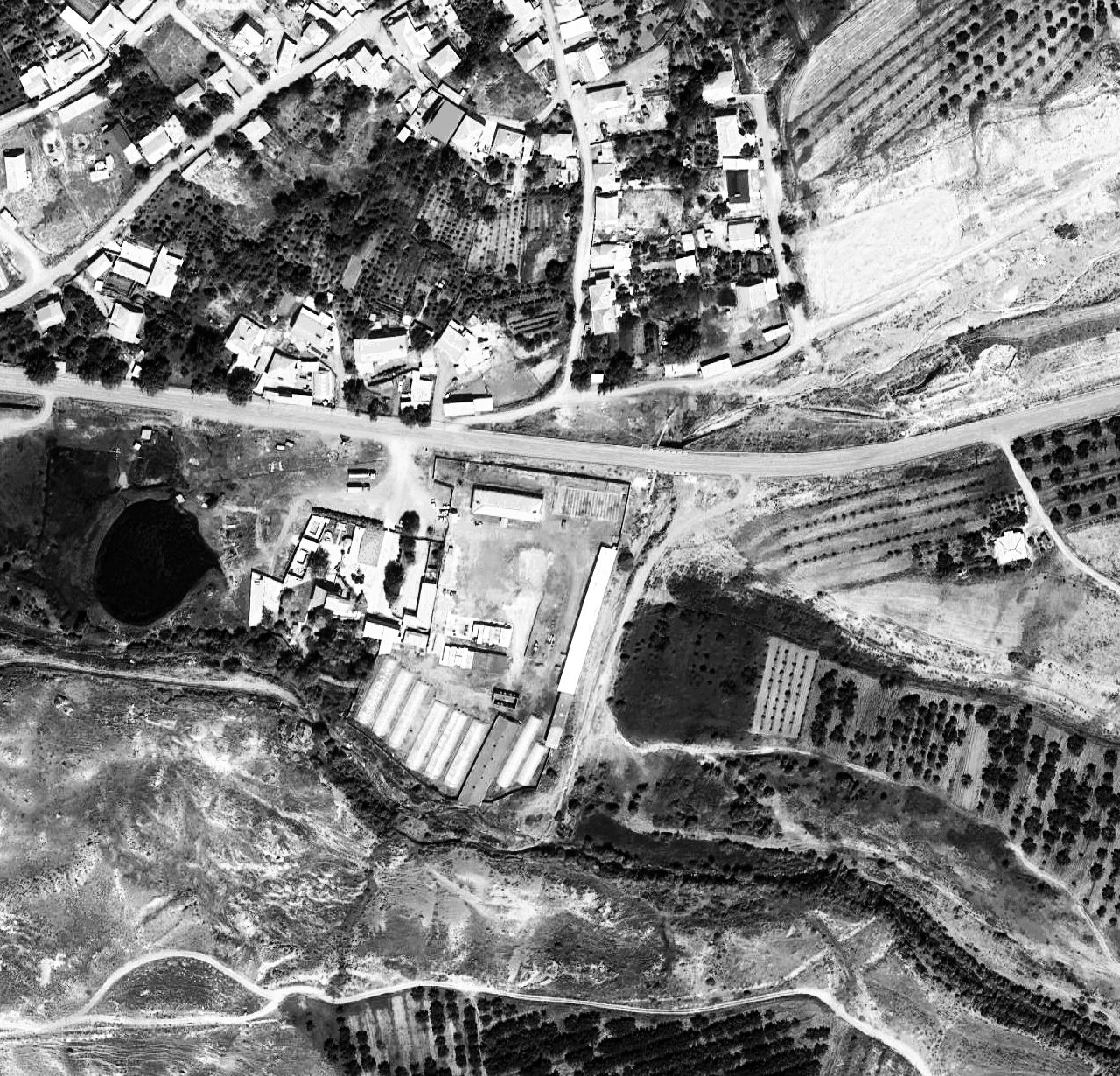}
    \end{minipage}
    % \vspace{0.01\textwidth}
    \begin{minipage}[b]{0.32\textwidth}
        \includegraphics[width=\textwidth]{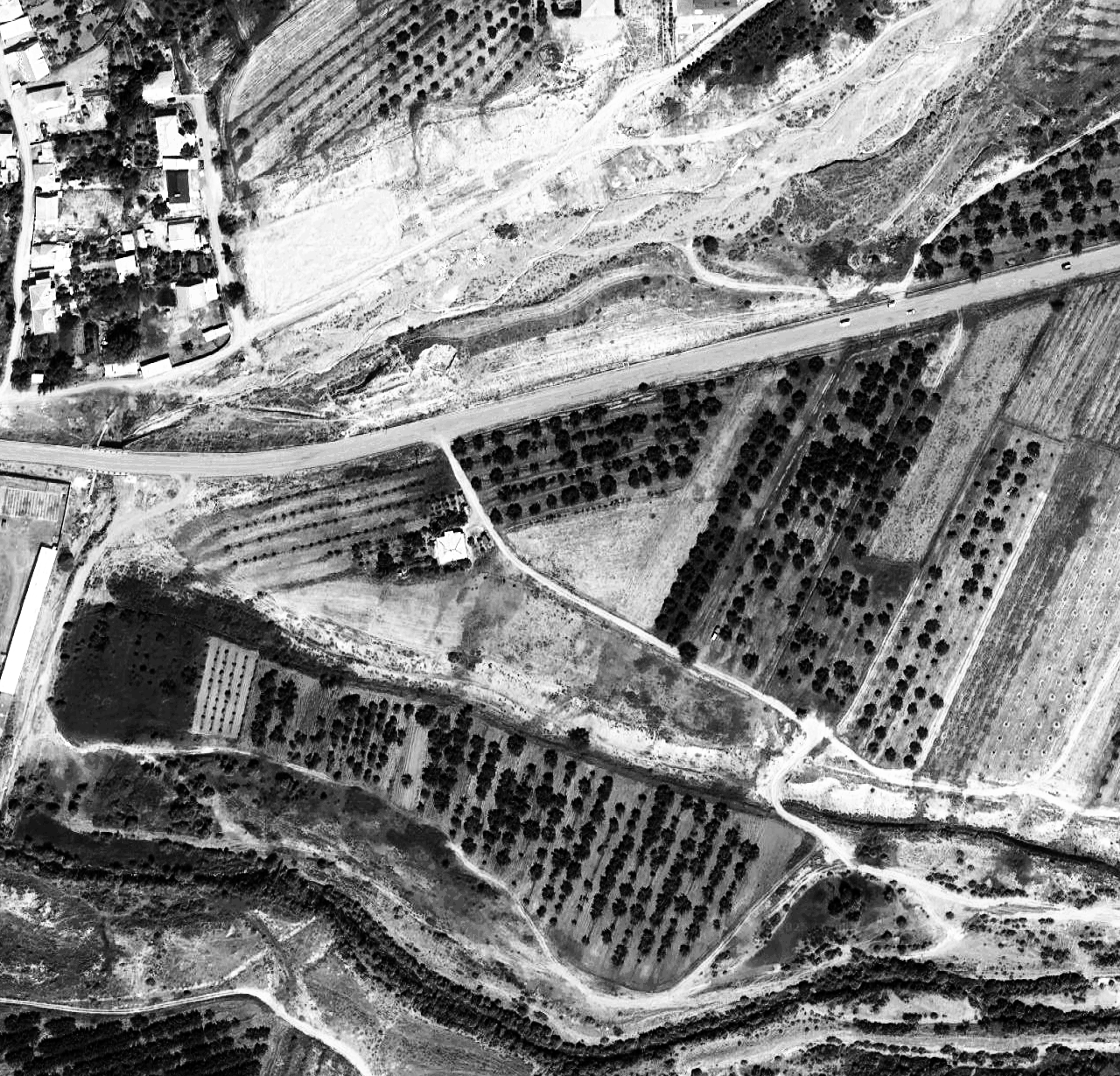}
    \end{minipage}

    \caption{Satellite Image Dataset. 
    The images are sampled such that the neighbouring images have overlapping regions, enable evaluation of the image matching and alignment algorithms.
    A collection of the satellite images is retrieved by utilizing the Google Maps Static API~\cite{gmaps_api}.}
    \label{fig:dataset_google}
\end{figure}

% Dataset Overview
To evaluate and benchmark the classical image matching techniques, we constructed a custom dataset (Figure~\ref{fig:dataset_google}) consisting of a satellite imagery collected by utilizing the Google Maps Static API~\cite{gmaps_api}, capturing images tile-by-tile over a predefined geographic region.
Each image is defined by the GPS coordinates of its center point and a fixed zoom level ensuring a consistent spatial resolution, scale and minimal distortion across the dataset.
The sampling grid was designed to ensure a partial overlap between the neigh\-bour\-ing images enabling a robust pairwise matching evaluation. 

% The dataset includes a variety of terrains, such as residential areas, industrial zones, and open landscapes, to reflect the variability commonly encountered in real-world geospatial tasks. 
% The dataset is designed to reflect realistic conditions, including urban, rural, and mixed terrains, and supports pairwise image matching with known spatial relationships.

% Objective
The present work aims to provide a rigorous analysis of the local image matching techniques
with a particular focus on classical local feature-based descriptors such as the Scale-Invariant Feature Transform (SIFT)~\cite{lowe2004distinctive} and the Oriented FAST and Rotated BRIEF (ORB)~\cite{rublee2011orb}
that is built on the Features from Accelerated Segment Test (FAST) keypoint detector~\cite{rosten2005fusing} and the Binary Robust Independent Elementary Features (BRIEF) descriptor~\cite{calonder2010brief}.
We also consider their integration with matching algorithms and geometric verification pipelines, such as the Random Sample Consensus (RANSAC)~\cite{fischler1981random},
as well as deliver a qualitative evaluation of their accuracy in the satellite imagery domain.

% Related work
A number of studies have explored the descriptors evaluation in various contexts. Miko\-lajczyk et al.~\cite{mikolajczyk2005performance} compared local descriptors under image transformations.
Panchal et al.~\cite{panchal2013comparison} analysed the SIFT and other algorithms based on their performance in the keypoint detection and descriptor matching.
Their evaluation conducted on a single dataset without cross-view variations, focused on the computation time, the number of detected keypoints, and the matching accuracy under affine and scale transformations.
Whereas the study provides valuable insights into the strengths and weaknesses of these classical methods in a controlled setting, it does not address more complex scenarios such as cross-view which are crucial for geo-localization and place recognition tasks.
Benchmarks such as in~\cite{balntas2017hpatches} provide the standardized evaluation frameworks.
Recent comparative studies~\cite{tareen2018comparative} have extensively benchmarked handcrafted de\-scrip\-tors on standard datasets highlighting their trade-offs in terms of the accuracy and com\-pu\-ta\-tional cost. 

The main contributions of the present paper are as follows:

\begin{itemize}
    \item We formulate the mathematical foundations underlying classical handcrafted keypoint descriptors including the SIFT and ORB.

    \item We evaluate the accuracy for image matching pipelines based on the RANSAC inliers including the Inlier Ratio metric.

    \item We conduct a systematic numerical comparison of local descriptors analyzing their performance under different accuracy metrics.
\end{itemize}

% Paper Organization
The paper structure is organized in the following way:
Section~2 introduces math\-e\-mati\-cal formulation of the digital image and notations adopted throughout the text.
Section~3 defines the concept of the keypoint and local descriptor formulation, provides an overview of the SIFT and ORB methods emphasizing their properties and differences.
Section~4 describes the keypoint matching procedure based on the descriptor similarity and presents the RANSAC algorithm for the geometric outlier rejection.
Section~5 outlines the accuracy metrics used to evaluate the keypoint matching quality.
Section~6 presents a numerical evaluation of the discussed methods on a benchmark dataset with analysis of the matching performance.
In Section~7, the summary is provided and future research directions are outlined.

\section{Image Definition}

The image can be mathematically defined as a function,
\[
I: \Omega \subset \mathbb{R}^2 \rightarrow \mathbb{R}^c,
\]
\noindent where the continuous domain \( \Omega \) represents an image in the OXY-plane but the range \( \mathbb{R}^c \) represents an intensity value with \( c \) channels (e.g. \( c = 1 \) for the grayscale image, \( c = 3 \) for the RGB image).
In the discrete case, the digital image is modeled as
\[
I: \mathcal{D} \subset \mathbb{Z}^2 \rightarrow \mathbb{R}^c,
\]
\noindent where the discrete domain \( \mathcal{D} = \{ 0, \dots, H-1 \} \times \{ 0, \dots, W-1 \} \) is the pixel locations grid,
$W > 0$ and $H > 0$ are the width and height of the image in pixels, respectively,
the origin \( O=(0,0) \) is the top-left pixel,
\( I(x, y) \in \mathbb{R}^c \) returns the pixel value (intensity or color vector) at the integer location $(x, y)$.

% Optionally, the values might be:
%     bounded, e.g., [0,255]⊂Z[0,255]⊂Z for 8-bit images
%     normalized, e.g., [0,1]⊂R[0,1]⊂R for float images
% To convert a color image \( I(x, y) \) from the RGB color space to a grayscale image \( I_{\text{gray}}(x, y) \), a weighted sum of the red, green, and blue color channels is computed:

\section{Scale-Invariant Feature Transform (SIFT)}

The keypoint (also called the interest point or feature point) is a distinctive and repeatable location in an image that can be used for matching points between images. Desirable properties of the ideal keypoint are repeatability (the same keypoint should be found in multiple views of the same scene).
It should have a unique distinctive local neighborhood and be invariant to scale changes, rotation, illumination and viewpoint transformations.

Early primitive keypoint detectors were focusing on finding locally distinctive points like corners or blobs without a full invariance to the scale or rotation.
One of the earliest and most influential methods is the Harris corner detector~\cite{harris1988combined}, which identifies points where the image intensity strongly changes in two directions -- typically at corners.
The detection exploits the Moravec operator~\cite{moravec1980obstacle}, which detects interest points based on shifting the windows in multiple directions.
Another early approach is the Laplacian of Gaussian~\cite{lindeberg1998feature}, which detects blob-like structures by locating the extrema in a scale-normalized Laplacian.
These detectors provide repeatable and distinctive keypoints but lack the scale or rotation invariance -- making them less robust under transformations compared to later methods like the SIFT.

The SIFT descriptor encodes local appearance of an image region surrounding a keypoint in a manner that is invariant to the scale, rotation, and moderate affine distortions. The descriptor is constructed through the following steps:

\paragraph{Scale-Space Construction:}
In the SIFT, the Gaussian function with the kernel
\[
G(x, y, \sigma) = \frac{1}{2\pi \sigma^2} \exp\left(-\frac{x^2 + y^2}{2\sigma^2} \right)
\]
\noindent is adopted, where the standard deviation \(\sigma\) controls amount of blurring and plays central role in keypoints detection.
The Gaussian blur smooths the image and reduces the high-frequency noise, thereby improving the stability of the keypoint detection.

A scale-space representation is created by convolving the image with the Gaussian kernels at different scales resulting in multi-scale image pyramid.
The continuous con\-vo\-lu\-tion in the scale-space construction is defined by
\[
L(x, y, \sigma) = \iint_{\mathbb{R}^2} I(\xi, \eta) \, G(x - \xi, y - \eta, \sigma) \, d\xi \, d\eta,
\]
\noindent where the variables \(\xi\) and \(\eta\) represent coordinates in a neighborhood of the point \((x, y)\).
They are used to compute the weighted sum of the image values by the Gaussian function.
Pixels closer to the center \( (x, y) \) contribute more due to the higher Gaussian values.
In the continuous space, the domain of integration is \(\mathbb{R}^2\), i.e., the entire image plane.
In discrete implementations, the convolution with the kernel size $k$ takes the form
\[
L(x, y, \sigma) = \sum_{i = -k/2}^{k/2} \sum_{j = -k/2}^{k/2} I(x - i, y - j) G(i, j, \sigma)
\]
or, simply,
\[
L(x, y, \sigma) = G(x, y, \sigma) * I(x, y),
\]
\noindent where $*$ is the convolution operator.

\begin{figure}[ht]
    \centering
    \includegraphics[width=0.95\textwidth]{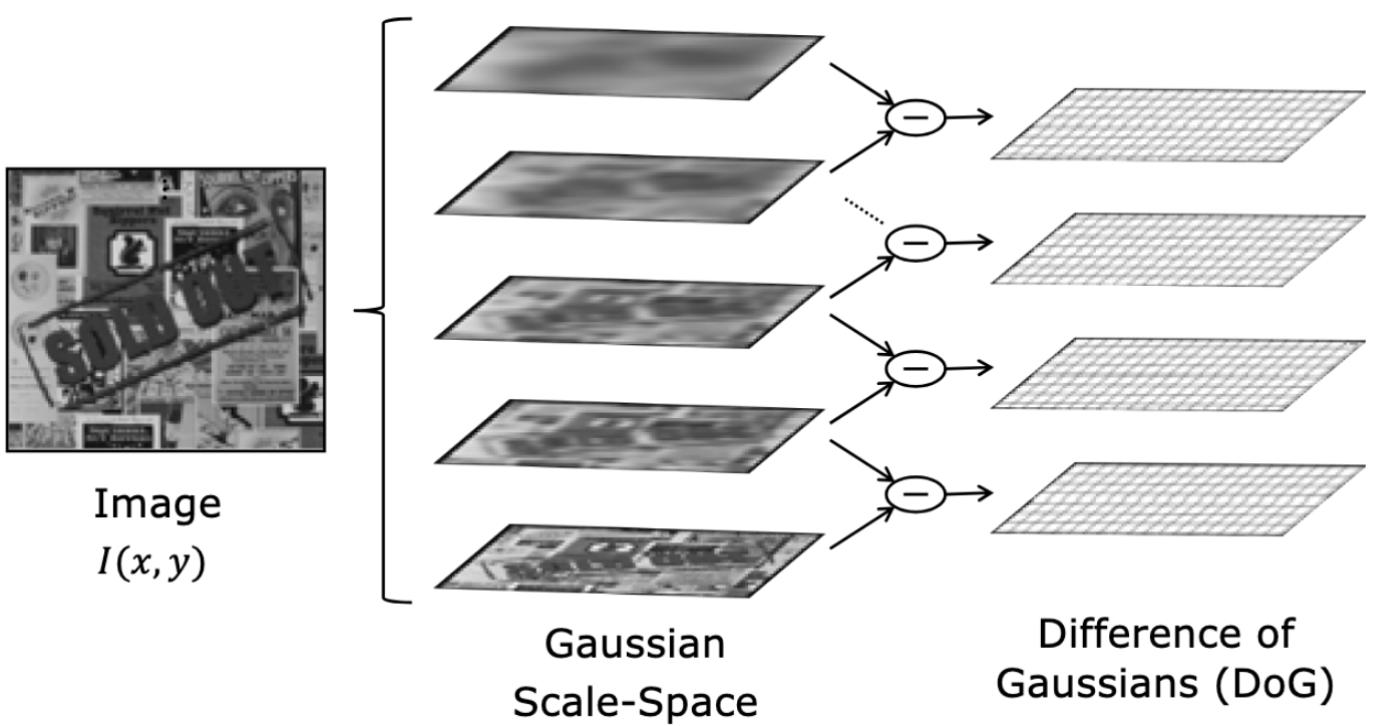}
    \caption{Difference of the Gaussians \cite{sift_slides}
    is computed by subtracting the scale-space representation of an input image at different levels of the Gaussian blurring resulting in a pyramid of images, which is utilized for detecting the scale-invariant keypoints.}
    \label{fig:dog}
\end{figure}

\paragraph{Keypoint Detection:}
Keypoints are identified as local extrema in the Difference of Gaussians function by repeatedly comparing each point with its neighbors across both spatial and scale dimensions, increasing the standard deviation \(\sigma\) producing different levels of blur (Figure~\ref{fig:dog}):
\[
D(x, y, \sigma) = L(x, y, k\sigma) - L(x, y, \sigma),
\]
\noindent where \(k\) is a multiplicative factor, typically \(\sqrt{2}\).

In the SIFT, a keypoint is considered if a pixel in the Difference of Gaussians image pyramid is a local extremum with respect to its neighborhood in the scale and space, that is, its value is either greater or less than all the surrounding pixels in both spatial and scale dimensions.

\paragraph{Orientation Assignment:}

The orientation assignment step in the SIFT is crucial to achieve rotation invariance of keypoints.
To make descriptors invariant to the image rotation, each keypoint is assigned to a dominant orientation based on the local image gradients.
The descriptor is then computed relative to this orientation so that the image rotation does not affect the descriptor. This means, no matter how the image is rotated, the descriptor will look in the same way.

For each keypoint, an orientation histogram is constructed from the gradient mag\-ni\-tudes and orientations within a local neighborhood.
The gradient magnitude and direction at each pixel are computed by
\[
m(x, y) = \sqrt{L_x^2 + L_y^2}
\quad \text{and} \quad
\theta(x, y) = \arctan\left( \frac{L_y}{L_x} \right),
\]
\noindent respectively, where \( L_x \) and \( L_y \) denote the image derivatives in the \( x \) and \( y \) directions, respectively, computed from the smoothed image \( L(x, y, \sigma) \).
The dominant orientation is selected as the histogram peak keeping the rotation invariance.

\paragraph{Descriptor Vector Construction:} % Local Histogram Computation

In the SIFT descriptor construction, a local image patch, which is centered at a detected keypoint, is extracted and processed to capture distinctive gradient-based information.
Around each detected keypoint $(x_p, y_p)$, a square patch $\Omega_p$ of the size $S \times S$ is considered.
The patch $\Omega_p$ is divided into the $M \times M$ spatial subregions of the size $s \times s$, where $s = S/M$.
In each subregion, an orientation histogram is constructed.
Denote the bin number as $B$ covering the angular range $[0, 2\pi)$ with the bin centers
\[
\theta_b = \frac{2\pi b}{B}, \quad b=0,1,\dots,B-1.
\]

Each pixel $(x,y)$ within a patch contributes to the histogram with a vote weighted by its gradient magnitude and the spatial Gaussian weighting function
\[
w(x,y) = \exp \left( - \frac{ (x - x_p)^2 + (y - y_p)^2 }{ 2\sigma^2 } \right).
\]

The orientation histogram for the subregion $r$ is as follows:
\[
H^r_b = \sum_{(x,y) \in r}{ w(x,y) \cdot m(x,y) \cdot \mathbbm{1}_b( \theta(x,y)) },
\quad
b = 1, 2, \dots, B,
\]
\noindent where $m(x, y)$ is the gradient magnitude, $\mathbbm{1}_b(\theta)$ is an indicator of assigning the gradient orientation to the bin $b$,
\[
\mathbbm{1}_b(\theta) = 
\begin{cases}
1, &  \theta \in \left[ \frac{2\pi(b-1)}{B} , \frac{2\pi b}{B} \right) \\   
0, & \text{otherwise}.
\end{cases} 
\]

After computing the histograms for all $M \times M$ subregions, they are concatenated into the single descriptor vector
\begin{equation} \label{eq:sift_descriptor}
    \mathbf{d}_\text{SIFT} = \left[ H^1_0, \dots, H^1_{B-1}, H^2_0, \dots, H^{M^2}_{B-1} \right].
\end{equation}
Thus, the total dimensionality of the SIFT descriptor is \( D = M^2 \times B\).
In practice, the SIFT descriptor is commonly constructed using a local image patch of the size \( 16 \times 16 \) pixels centered at the keypoint.
This patch is divided into a grid of \( 4 \times 4 \) spatial subregions, and each subregion contributes a histogram of gradient orientations.
Each orientation histogram typically contains 8 bins.
As a result, the final descriptor is a concatenation of \( 4 \times 4 \) histograms, each with 8 bins yielding a feature vector of the dimension $D = 4^2 \times 8 = 128$.

% \paragraph{Normalization and Clipping:}

\section{Oriented FAST and Rotated BRIEF (ORB)}

The ORB descriptor is a fast and efficient alternative to the SIFT combining a keypoint detector based on the FAST algorithm with an orientation component and a binary descriptor derived from BRIEF.
It is particularly suitable for real-time applications and devices with limited computational resources.

\paragraph{Keypoint Detection using the FAST:}

\begin{figure}[ht]
    \centering
    \includegraphics[width=0.8\textwidth]{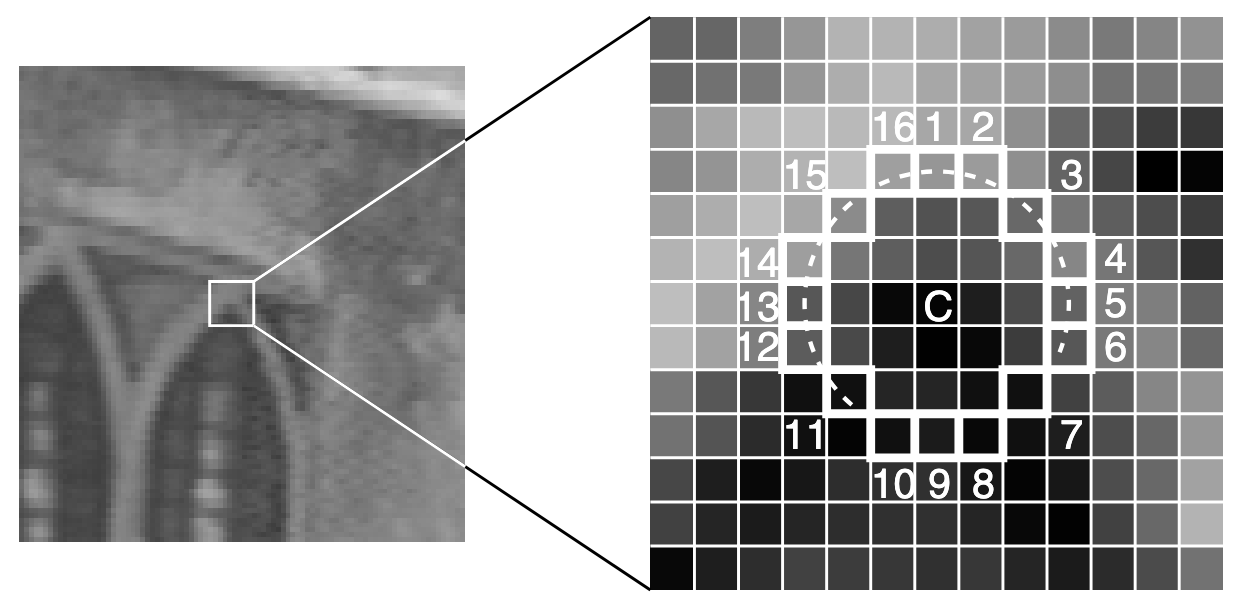}
    \caption{The FAST feature detection \cite{rosten2005fusing}.
    As example, the Bresenham circle of the radius 3 with the center at $C$ is presented.
    The highlighted squares are the pixels adopted in the feature detection.}
    \label{fig:fast}
\end{figure}

The ORB employs the FAST detector, which performs binary intensity tests around a circular region.
A point $(x, y)$ is a FAST keypoint if there exists a contiguous arc of, at least, $N$ pixels in the Bresenham circle centered at $(x, y)$ (Figure~\ref{fig:fast})
that are either significantly brighter or darker than the center point, i.e.,
\[
I(x_q,y_q) > I(x,y) + t
\quad \text{or} \quad
I(x_q,y_q) < I(x,y) - t,
\quad q = 1, 2, \dots, B,
\]
\noindent where $t$ is an intensity threshold,
${(x_q,y_q)}_{q=1}^B$ is the set of $B$ points lying on the Bresenham circle centered at $(x, y)$.
The use of a circular patch is important because it provides rotational symmetry around the candidate pixel ensuring that the detection is invariant to image rotations.
This results in the high-speed feature detection but without a strength score.

To rank and filter the FAST keypoints, the ORB computes the Harris corner measure \cite{harris1988combined} by using the second moment matrix

\[
M = 
\begin{bmatrix}
\sum I_x^2 & \sum I_x I_y \\
\sum I_x I_y & \sum I_y^2
\end{bmatrix}, \quad 
\\
H(x,y) = \det(M) - \alpha \cdot (\text{trace}(M))^2,
\]
\noindent where $I_x$ and $I_y$ denote the image derivatives in horizontal and vertical directions respectfully; $\alpha$ is a sensitivity constant, typically, $\alpha \in [0.04, 0.06]$.

% \paragraph{Orientation Assignment:}
% FAST itself does not provide an orientation for the keypoints it detects -- it just finds corners.
% ORB enhances FAST by computing orientation using the intensity centroid method:

% \[
% \theta = \arctan\left( \frac{m_{01}}{m_{10}} \right), \quad m_{pq} = \sum_x \sum_y x^p y^q I(x, y)
% \]
% This adds rotational invariance to the FAST keypoints.

% To ensure rotation invariance, ORB computes an orientation $\theta$ for each keypoint using the intensity centroid method. Define the moments:

% \[
% m_{pq} = \sum_{x, y} x^p y^q I(x, y)
% \]

% Then the centroid is:

% \[
% \mathbf{C} = \left( \frac{m_{10}}{m_{00}}, \frac{m_{01}}{m_{00}} \right)
% \]

% and the orientation is:

% \[
% \theta = \arctan\left( \frac{m_{01}}{m_{10}} \right)
% \]

\paragraph{Rotated BRIEF Descriptor:}

The ORB adopts the BRIEF descriptor, which relies on intensity comparisons between the point pairs. 
Typically, the pairs are drawn ran\-domly according to an Gaussian distribution centered at the keypoint with zero mean and standard deviation proportional to the patch size.
This strategy ensures that the pairs are well distributed spatially and do not cluster only near the center. 
Let a local patch \(\Omega\) centered at a keypoint be defined on a square domain of the size \( S \times S \) pixels.
In the BRIEF descriptor, we sample \( N \) pairs of the pixel coordinates
\[
\left\{ \left( \mathbf{p}_i, \mathbf{q}_i \right) \,\middle|\, \mathbf{p}_i, \mathbf{q}_i \in \Omega \subset \mathbb{R}^2 \right\}_{i=1}^N.
\]

Each pair defines a binary intensity test,
\[
\tau_i = 
\begin{cases}
1, & I(\mathbf{p}_i) < I(\mathbf{q}_i) \\
0, & \text{otherwise}.
\end{cases}
\]

Repeating this test for \( N \) pairs yields the \( N \)-bit binary BRIEF descriptor
\begin{equation} \label{eq:brief_descriptor}
    \mathbf{d}_\text{BRIEF} = [\tau_1, \tau_2, \dots, \tau_N] \in \{0,1\}^N.
\end{equation}

To achieve the rotation invariance, the test pairs are rotated by the dominant orientation angle $\theta \in [0, 2\pi)$ which is selected by computing a histogram of gradient orientations within the image patch around the keypoint
\[
\mathbf{p}_i' = \mathbf{R}_\theta \mathbf{p}_i, \quad \mathbf{q}_i' = \mathbf{R}_\theta \mathbf{q}_i, \quad
\mathbf{R}_\theta = 
\begin{bmatrix}
\cos \theta & -\sin \theta \\
\sin \theta & \cos \theta
\end{bmatrix},
\]
\noindent where \( \mathbf{R}_\theta \) is the rotation matrix,
\( (\mathbf{p}_i', \mathbf{q}_i') \) are the resulting rotated coordinates to ensure the rotation invariance.

\section{Keypoint Matching Techniques}

Image matching is the task of establishing visual correspondences between two images by comparing local features.
The standard pipeline consists of detecting the keypoints, extracting the local descriptors, matching these descriptors, and verifying them by using geometric constraints.
To compare the descriptors extracted from images, we employ standard similarity measures.

\paragraph{SIFT Descriptor Matching:}
Commonly used metric for the SIFT is the Euclidean distance (also called \(L_2\)-norm). Euclidean distance measures the straight-line distance between two descriptors.
Given two descriptors \( \mathbf{d}_1, \mathbf{d}_2 \in \mathbb{R}^D \), the Euclidean distance is defined as
\begin{equation} \label{eq:euclidean}
d_\text{Euclidean}(\mathbf{d}_1, \mathbf{d}_2) = \| \mathbf{d}_1 - \mathbf{d}_2 \|_2 = \sqrt{\sum_{i=1}^{D} (\mathbf{d}_{1,i} - \mathbf{d}_{2,i})^2},
\end{equation}
\noindent where $\mathbf{d}$ is the SIFT descriptor defined in~\eqref{eq:sift_descriptor} and $D$ is the descriptor dimensionality.

\paragraph{BRIEF Descriptor Matching:}
The BRIEF descriptors are matched by utilizing the Hamming distance.
The Hamming distance is defined as the number of positions at which their bits differ, i.e.,
\begin{equation} \label{eq:hamming}
d_\text{Hamming}(\mathbf{d}_1, \mathbf{d}_2) = \sum_{i=1}^N \mathbbm{1} \left[ \mathbf{d}_{1,i} \neq \mathbf{d}_{2,i} \right],
\end{equation}
\noindent where $\mathbf{d}$ is the BRIEF descriptor defined in~\eqref{eq:brief_descriptor},
$N$ is the descriptor dimensionality
and $\mathbbm{1}[\cdot]$ indicates whether the $i$-th bits of the binary descriptor vectors $\mathbf{d}_1$ and $\mathbf{d}_2$ differ.
\noindent In other words, the Hamming distance is simply the count of the bit mismatches between the two binary strings, making it computationally efficient for comparing the BRIEF descriptors.
% Equivalently, using the bitwise exclusive-or (XOR) operation:

\paragraph{Descriptor Matching with Brute Force:}

Let \( \{x_i\}_{i=1}^{N} \subset \mathbb{R}^D \) and \( \{y_j\}_{j=1}^{M} \subset \mathbb{R}^D \) be the sets of descriptors extracted from the first and second image, respectively, where \( D \) is the descriptor dimension.
The Brute Force Matcher compares each descriptor \( x_i \) with every \( y_j \) and finds the best match \( j^* \) for each \( i \) by minimizing a distance metric \( d \):
\[
j^* = \arg\min_j d(\mathbf{d}_i, \mathbf{d}_j).
\]
\noindent where distance metric is $d_\text{Euclidean}$~\eqref{eq:euclidean} for the SIFT and $d_\text{Hamming}$~\eqref{eq:hamming} for the ORB.
We define the set of all the Brute Force matches as
\begin{equation} \label{eq:matches_bf}
    \mathcal{M} = \left\{ (\mathbf{x}_i, \mathbf{x}_j) \mid \mathbf{x}_j = \arg\min_{j} d(\mathbf{d}_i, \mathbf{d}_j) \right\}.
\end{equation}

\begin{figure}[ht]
    \centering
    \includegraphics[width=0.95\textwidth]{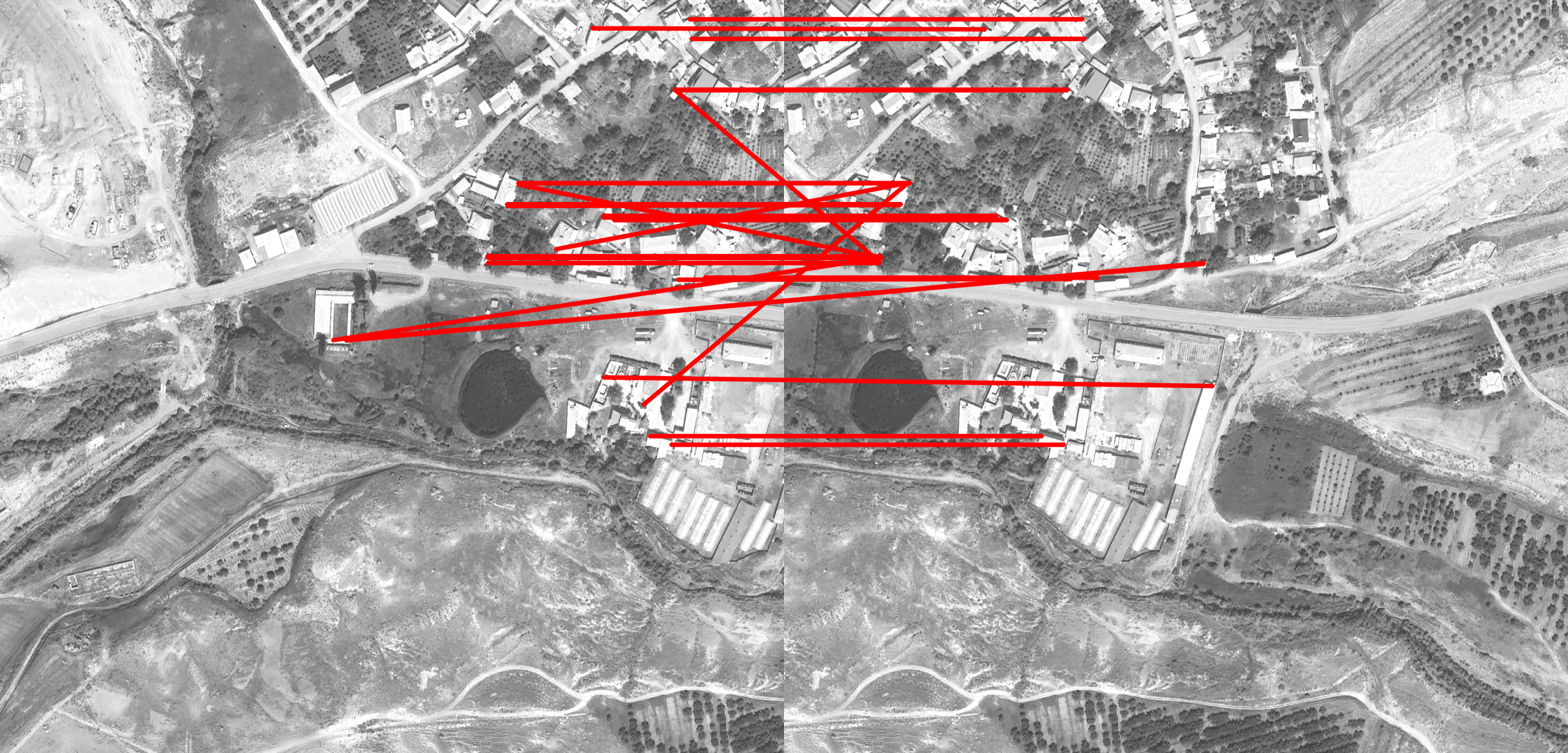}
    \caption{Keypoint matches with the Brute Force.
    Initial keypoint correspondences between two images using a Brute Force Matcher on SIFT descriptors, including outliers.
    Images are retrieved by utilizing the Google Maps Static API~\cite{gmaps_api}.}
    \label{fig:matches_bf}
\end{figure}

Many incorrect matches (outliers) are present due to the lack of geometric ver\-ifi\-cation (Figure~\ref{fig:matches_bf}).

\paragraph{Homography Estimation:}
A homography is a projective transformation that maps points from one image plane to another under perspective geometry.
It is represented by a \( 3 \times 3 \) matrix \( \mathbf{H} \in \mathbb{R}^{3 \times 3} \), and is defined up to a non-zero scale factor.
Given a point \(\mathbf{x} = (x, y, 1)\) in the homogeneous coordinates, the homography relates it to its image \( \mathbf{x}' = (x', y', 1)\) in the second view as
\[
\lambda \mathbf{x'} = \mathbf{H} \mathbf{x}
\]

\noindent where \(\lambda\) is a nonzero scale factor (due to homogeneous coordinates) and
\[
\mathbf{H} =
\begin{bmatrix}
h_{11} & h_{12} & h_{13} \\
h_{21} & h_{22} & h_{23} \\
h_{31} & h_{32} & h_{33}
\end{bmatrix}
\]
\noindent is the parameterized homography with elements $h_{ij}$ that should be found from a set of corresponding points between two images.

For each point correspondence \( (x, y) \leftrightarrow (x', y') \), the projective mapping yields
\[
\begin{aligned}
x'(h_{31}x + h_{32}y + h_{33}) &= h_{11}x + h_{12}y + h_{13}, \\
y'(h_{31}x + h_{32}y + h_{33}) &= h_{21}x + h_{22}y + h_{23}.
\end{aligned}
\]

Rewriting the above as linear equations in the unknown vector \( \mathbf{h} \in \mathbb{R}^9 \), where
\[
\mathbf{h} = 
\begin{bmatrix}
h_{11} & h_{12} & h_{13} & h_{21} & h_{22} & h_{23} & h_{31} & h_{32} & h_{33}
\end{bmatrix}^\top,
\]
\noindent we obtain the following linear system for each correspondence:

\[
\begin{bmatrix}
x & y & 1 & 0 & 0 & 0 & -x x' & -y x' & -x' \\
0 & 0 & 0 & x & y & 1 & -x y' & -y y' & -y'
\end{bmatrix}
\cdot
\mathbf{h}
= \mathbf{0}.
\]

The matrix \( \mathbf{H} \) has 8 degrees of freedom. It is usually normalized by assuming \(h_{33} = 1\).
Stacking these equations for \( n \geq 4 \) point correspondences yields the linear system
\[
A \mathbf{h} = \mathbf{0},
\quad A \in \mathbb{R}^{2n \times 9}.
\]

Solving the system provides the elements $h_{ij}$, which are then reshaped into the homography matrix $\mathbf{H}$.

% \section{Solving via SVD}

\paragraph{Geometric Verification with the RANSAC:}

In practice, point correspondences often include outliers due to mismatches.
To ensure spatial consistency, the matched keypoints are verified by estimating a geometric transformation. 
To estimate \( \mathbf{H} \) robustly in the presence of outliers, the RANSAC algorithm is adopted.
It randomly samples subsets of matches, computes candidate homographies, and selects the one that maximizes the number of inliers -- matches that satisfy

\[
\| \mathbf{x}'_i - \mathbf{H} \mathbf{x}_i\| < \epsilon
\]
\noindent for a predefined threshold \( \epsilon \).
Repeat the above steps for a fixed number of iterations or until a satisfactory consensus is found.
This approach effectively minimizes the influence of outliers and improves the matching accuracy.

\paragraph{Final Inlier Matches:}

\begin{figure}[ht]
    \centering
    \includegraphics[width=0.95\textwidth]{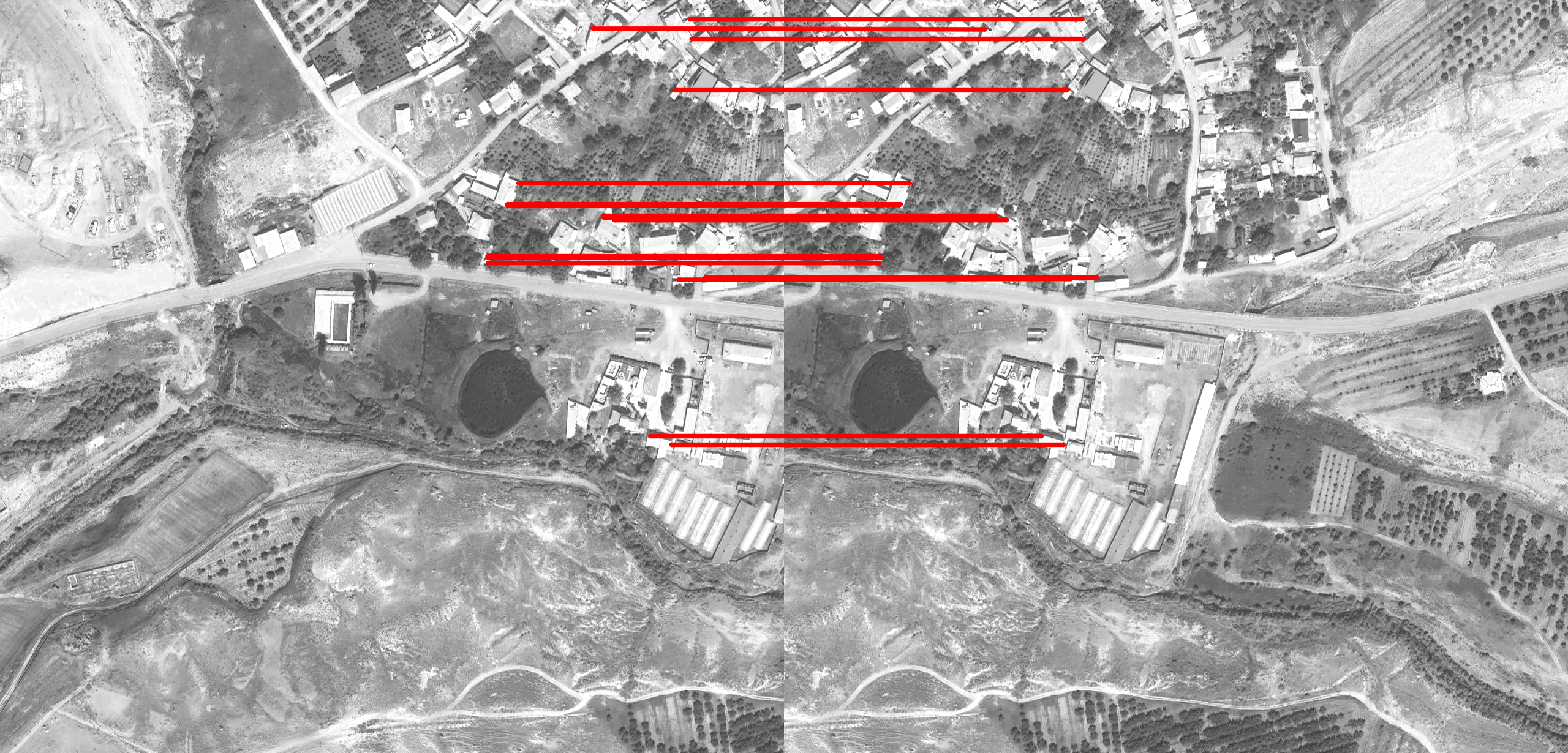}
    \caption{The keypoint matches between two images after applying the RANSAC filtering.
    The RANSAC rejects outliers and retains only geometrically consistent matches (inliers),
    leading to a much cleaner and structurally meaningful correspondence set.
    Images are retrieved by utilizing the Google Maps Static API~\cite{gmaps_api}.}
    \label{fig:matches_ransac}
\end{figure}

The final set of matches taken for the image alignment or stitching includes only whose are consistent with the estimated homography,
\begin{equation} \label{eq:matches_inlier}
    \mathcal{M}_{\text{Inlier}} = \{ (\mathbf{x}_i, \mathbf{x}'_i) \mid \|\mathbf{x}'_i - \mathbf{H} \mathbf{x}_i\| < \epsilon \},
\end{equation}
\noindent where \( \epsilon \) is the reprojection error threshold. It specifies the maximum allowed deviation between the projected point \( \mathbf{H} \mathbf{x}_i \) and the observed point \( \mathbf{x}'_i \) for the correspondence to be classified as an inlier.
Smaller values of \( \epsilon \) yield stricter geometric consistency at the cost of rejecting more matches.
This geometric verification step significantly reduces false matches and improves the reliability of downstream tasks such as image registration and structure-from-motion (Figure~\ref{fig:matches_ransac}).

\section{Accuracy Metrics}

In the image matching pipelines based on local descriptors and geo\-met\-ric ver\-i\-fi\-ca\-tion, one common approach to assess correctness is through the analy\-sis of RANSAC in\-liers -- matches that conform to a geometric model under a reprojection error threshold.

\paragraph{Geometric Inlier Definition:}

Let \( \mathcal{M}_\text{Inlier}(I_i,I_j) \subset \mathcal{M}(I_i,I_j) \) denote the set of inliers between images $I_i$ and $I_j$,
where $\mathcal{M}$ is defined in \eqref{eq:matches_bf} and $\mathcal{M}_\text{Inlier}$ is defined in \eqref{eq:matches_inlier}.
The following metrics can be used to quantify the match quality:

\paragraph{Inlier Ratio:} Computes the fraction of total matches that are inliers,
\[
R_{\text{Inlier}} = \frac{|\mathcal{M}_\text{Inlier}|}{|\mathcal{M}|},
\]
\noindent where $\mathcal{M}$ and $\mathcal{M}_\text{Inlier}$ are defined in \eqref{eq:matches_bf} and \eqref{eq:matches_inlier}, respectively.

% \[
% |M_{\text{Inlier}|} \geq \theta, \quad
% R_{\text{Inlier}} \geq \rho,
% \]
% where \( \theta \in \mathbb{N} \) is a minimum inlier count threshold, and 
% These thresholds may be tuned depending on dataset properties and application requirements.

\paragraph{Matching Decision Rule:}
Define a Prediction function $\mathcal{P}$ with threshold $\rho$ com\-puted for the given pair of images by using the matching pipeline,
\begin{equation} \label{eq:prediction_f}
    \mathcal{P}_{\rho}(I_i, I_j) =
    \begin{cases}
    1, & \text{if } R_{\text{Inlier}}(I_i, I_j) \geq \rho \\
    0, & \text{otherwise},
    \end{cases}
\end{equation}

\noindent where \( \rho \in (0, 1) \) is a the minimum Inlier Ratio.
An image pair is considered a correct match if $\mathcal{P}_{\rho}(I_i, I_j) = 1$.

\paragraph{Ground truth:}
The evaluation of the feature matching performance relies on com\-par\-ing the predicted matches to the true relationships between images.
The ground truth refers to the known, correct match status of each image pair in the dataset.
In the context of visual place recognition on the map images, the ground truth defines whether two images represent the same location (match) or different locations (non-match).
These labels are typically obtained from the manual annotation, spatial metadata, or reliable prior mapping,
\begin{equation} \label{eq:gt_f}
    \mathcal{G}(I_i, I_j) =
    \begin{cases}
    1, & I_i \text{ and } I_j \text{ represent the same location}, \\
    0, & \text{otherwise}.
    \end{cases}
\end{equation}

Based on the Ground Truth and Prediction functions, each matched image pair is categorized in the following way:
\begin{align*}
\text{True Positive: TP} &= \sum_{i,j} \mathbbm{1} \left[ \mathcal{G}(I_i, I_j) = 1 \land \mathcal{P}_{\rho}(I_i, I_j) = 1 \right], \\
\text{True Negative: TN} &= \sum_{i,j} \mathbbm{1} \left[ \mathcal{G}(I_i, I_j) = 0 \land \mathcal{P}_{\rho}(I_i, I_j) = 0 \right], \\
\text{False Positive: FP} &= \sum_{i,j} \mathbbm{1} \left[ \mathcal{G}(I_i, I_j) = 0 \land \mathcal{P}_{\rho}(I_i, I_j) = 1 \right], \\
\text{False Negative: FN} &= \sum_{i,j} \mathbbm{1} \left[ \mathcal{G}(I_i, I_j) = 1 \land \mathcal{P}_{\rho}(I_i, I_j) = 0 \right],
\end{align*}
\noindent where $\mathcal{P}$ and $\mathcal{G}$ are defined in~\eqref{eq:prediction_f} and~\eqref{eq:gt_f}, respectively and 
$\mathbbm{1}[\cdot]$ indicates if the logical expression in brackets is true.

% \begin{itemize}
%     \item True Positive (TP): a pair correctly predicted as a match, in agreement with the ground truth.
%     \item True Negative (TN): a pair correctly predicted as a non-match, in agreement with the ground truth.
%     \item False Positive (FP): a pair incorrectly predicted as a match, while the ground truth states they are a non-match (i.e., a false alarm).
%     \item False Negative (FN): a pair incorrectly predicted as a non-match, while they are in fact a match (i.e., a missed detection).
% \end{itemize}

\section{Comparative Study}

% \section{Implementation Details}
% Describe the software environment, libraries, and parameter settings.

To evaluate the performance of local feature descriptors for the image matching, we conducted a systematic comparison of the SIFT and ORB based on the Inlier Ratio.
This metric provides a direct and interpretable measure of the matching reliability, especially in the absence of the ground truth keypoint correspondences.
Each image is compared one-by-one against the entire set of reference images. The ground truth (whether the two map images truly match the same place) provides the correct label.
A feature-based matching pipeline is applied consisting of the keypoint de\-tec\-tion, de\-scrip\-tor extraction, descriptor matching, and geometric verification using the RANSAC.
For each image pair, we estimate homography and count the number of inlier cor\-re\-spon\-dences as a measure of matching confidence.
We focus on Inlier Ratio alone and do not compute standard clas\-si\-fi\-ca\-tion metrics such as precision, recall, or $F_1$-score.
This is because the True Positives naturally exhibit high Inlier Ratios, while the True Negatives exhibit near-zero Inlier Ratios, making binary classification thresholds unnecessary.
Experiments were con\-ducted by varying the number of keypoints extracted per image: 100, 200, 500, 1000, and 2000.
For each configuration, the mean Inlier Ratio was computed across all image pairs in the dataset.
Both ORB and SIFT were tested under identical conditions (Table~\ref{tab:feature_costs}).
% Given the significant class imbalance -- with many more non-matching pairs than matching ones -- conventional accuracy is inadequate. Therefore, we evaluate the matching performance using the following metrics:

% Let \( \mathcal{I}_A = \{ I_1^A, I_2^A, \dots, I_N^A \} \) and \( \mathcal{I}_B = \{ I_1^B, I_2^B, \dots, I_N^B \} \) be subsets of two datasets described, each containing \( N \) images. Subsets are built in a way that geographic coverage of the images in every subset does not overlap and each image in \( \mathcal{I}_A \) has a known correct match in \( \mathcal{I}_B \), such that the ground truth correspondence is:

% To build the confusion matrix, we compute the number of inlier matches between each pair \( (I_i^A, I_j^B) \), where \( i, j \in \{1, \dots, N\} \).

% This indicates that the correct match (according to ground truth) has significantly more inliers than incorrect pairs.
% We can visualize \( \mathbf{M} \) as a heatmap to evaluate the recognition performance. The strength of the diagonal gives an intuitive indication of matching accuracy. % (Figure~\ref{fig:conf_matrix}).

\begin{center}
\begin{tabular}{|c|c|c|}
\hline

Algorithm
& Inlier Ratio (True Positives) 
& Inlier Ratio (True Negatives) \\

\hline
ORB (100)   & 24.94\% & 0.52\% \\
ORB (200)   & 26.48\% & 0.61\% \\
ORB (500)   & 27.77\% & 0.41\% \\
ORB (1000)  & 28.54\% & 0.23\% \\
ORB (2000)  & 29.34\% & 0.14\% \\
SIFT (100)  & 32.21\% & 0.58\% \\
SIFT (200)  & 32.81\% & 0.63\% \\
SIFT (500)  & 34.45\% & 0.43\% \\
SIFT (1000) & 35.29\% & 0.26\% \\
SIFT (2000) & 36.13\% & 0.15\% \\
\hline
\end{tabular}
\captionof{table}{Comparison of the SIFT and ORB methods based on the matching accuracy measured by inlier correspondences, in \%.
The number next to each method indicates the number of keypoints extracted per image.}
\label{tab:feature_costs}
\end{center}

The experimental results demonstrate that the SIFT consistently outperforms the ORB across all feature counts in terms of the Inlier Ratio.
Even when extracting only 100 keypoints, the SIFT achieves higher matching accuracy than the ORB at 2000 keypoints, emphasizing its robustness and suitability for the satellite image alignment tasks.
Moreover, the performance gains of increasing the features number are not significant after a certain threshold.
For instance, when using the SIFT, increasing the features number from 1000 to 2000 results in a noticeable improvement in Inlier Ratio from 35.29\% to 36.13\% -- less than 1\% gain, despite doubling the number of extracted keypoints.
The experiments suggest that extracting 200 to 500 keypoints strikes an effective trade-off between the accuracy and com\-pu\-ta\-tional cost, making these configurations particularly attractive for deployment on resource-constrained platforms,
while extracting more features may not be justified.
% Future work may explore dimensionality reduction techniques and learned descriptors to further improve matching quality without sacrificing efficiency.

\section{Conclusions}

A comparative analysis of classical feature matching algorithms applied to the satel\-lite imagery is presented.
A particular focus was placed on evaluation of the matching accuracy using the standard Computer Vision pipelines involving the keypoint detection, descriptor extraction, feature matching, and geometric verification through the RANSAC.
Quantitative evaluation was conducted by using the Inlier Ratio metric constructed from ground-truth correspondences between image pairs.
We provide a systematic framework for evaluating classical matching techniques.
The presented meth\-ods and findings can serve as a foundation for further studies of scenarios, where com\-pu\-ta\-tional and memory resources are limited.

In the future work, we aim to extend this analysis by
\begin{itemize}
    \item evaluating the cross-view matching between aerial and satellite imagery where scale, and lighting vary significantly,
    \item incorporating the learning-based methods such as the SuperPoint~\cite{superpoint} for im\-proved robustness,
    \item experimenting with descriptor aggregation techniques such as the VLAD~\cite{vlad} for efficient image retrieval.
\end{itemize}
Such extensions will further bridge the gap between classical vision techniques and modern learning-based systems in the context of the geospatial image analysis and localization.

\end{document}